\documentclass[journal]{IEEEtran}
\usepackage{cite}
\usepackage{amsmath,amssymb,amsfonts}
\usepackage{graphicx}
\usepackage{textcomp}
\usepackage{xcolor}
\usepackage{algorithm}
\usepackage{algorithmic}
\usepackage{multirow}
\usepackage{enumerate}
\usepackage{url}
\usepackage{booktabs}
\usepackage{threeparttable}
\usepackage{endnotes}
\usepackage{subfigure}
\usepackage{comment}
\usepackage[switch,pagewise,columnwise]{lineno}
\usepackage[colorlinks=true,linkcolor=blue,citecolor=blue,urlcolor=blue]{hyperref}
\newtheorem{def1}{\textbf{Definition}}

\ifCLASSINFOpdf
\else
\fi

\begin{document}
%
\title{Heuristic Search for Rank Aggregation with Application to Label Ranking}
\author{Yangming~Zhou,
        Jin-Kao Hao,
        Zhen~Li,
        and Fred~Glover
\thanks{This work was partially supported by the National Natural Science Foundation of China under Grant No. 61903144 and No. 72031007, and in part by the Shanghai Sailing Program under Grant No. 19YF1412400. (\emph{Corresponding author: Jin-Kao Hao})}
\thanks{Yangming~Zhou is with the Sino-US Global Logistics Institute, Shanghai Jiao Tong University, Shanghai 200030, China (e-mail: yangming.zhou@sjtu.edu.cn).}
\thanks{Jin-Kao~Hao is with the Department of Computer Science, Universit\'{e} d'Angers, Angers 49045, France (e-mail: jin-kao.hao@univ-angers.fr).}
\thanks{Zhen~Li is with the Tencent Technology (Shanghai) Co., Ltd, Shanghai 200233, China (email:acremanli@tencent.com).}
\thanks{Fred Glover is with the Leeds School of Business, University of Colorado, Boulder, CO 80309 USA (e-mail: glover@opttek.com).}

}


\maketitle
\begin{abstract}

Rank aggregation aims to combine the preference rankings of a number of alternatives from different voters into a single consensus ranking. As a useful model for a variety of practical applications, however, it is a computationally challenging problem. In this paper, we propose an effective hybrid evolutionary ranking algorithm to solve the rank aggregation problem with both complete and partial rankings. The algorithm features a semantic crossover based on concordant pairs and a late acceptance local search reinforced by an efficient incremental evaluation technique. Experiments are conducted to assess the algorithm, indicating a highly competitive performance on benchmark instances compared with state-of-the-art algorithms. To demonstrate its practical usefulness, the algorithm is applied to label ranking, which is an important machine learning task.

\end{abstract}


%
\IEEEpeerreviewmaketitle

\section{Introduction}
\label{Sec:Introduction}

Rank aggregation is a classical problem in voting theory, where each voter provides a preference ranking on a set of alternatives, and the system aggregates these rankings into a single consensus preference order to rank the alternatives. Rank aggregation plays a critical role in a variety of applications such as collaborative filtering \cite{Huang2011,Li2021}, multiagent planning \cite{Gharaei2021}, information retrieval \cite{Tamine2022}, and label ranking \cite{Destercke2015,Zhou2018,Dery2021}. As a result, this problem has been widely studied, particularly in social choice theory and artificial intelligence.

Given a set of $m$ labels $\mathcal{L}_m = \{\lambda_1,\lambda_2,\ldots,\lambda_m\}$, a ranking with respect to $\mathcal{L}_m$ is an ordering of all (or some) labels that represent an agent's preference for these labels. Rankings can be either complete or partial. A complete ranking includes all the labels and can be identified with a permutation $\pi$ of the set $\{1,2,\ldots,m\}$ such that $\pi(\lambda_i)$ denotes the position of $\lambda_i$ in the ranking $\pi$, that is, the rank of the label $\lambda_i$ in the ranking $\pi$.
For two labels $\lambda_i$ and $\lambda_j$, $\pi(\lambda_i) < \pi(\lambda_j)$ indicates that $\lambda_i$ is preferred over $\lambda_j$ and this preference relation is represented by $\lambda_i \prec \lambda_j$. However, real-world problems usually include partial rankings, where only $m'$ ($2 \leq m' < m$) labels are ranked. For example, when customers' preference relations about a set of movies, books, and laptops, are collected, the preference information on some labels may not be available. In this case, partial rankings can be used to express these partial preference relations.

Rankings can also be classified as with or without ties. A tie means there is no preference information among the ranked labels. The tied labels constitute a bucket. Therefore, an arbitrary ranking $\sigma$ can be represented as a list of its disjointed buckets, ordered from the most to the least preferred, and separated by vertical bars. The labels between two consecutive vertical bars indicate a bucket. For instance, let $\mathcal{L}_4 = \{\lambda_1,\lambda_2,\lambda_3,\lambda_4\}$ be the set of four labels. Then, $\sigma_1=(1|4|3|2)$ represents a complete ranking without ties; $\sigma_2=(1|3,4|2)$ represents a complete ranking with ties; $\sigma_3=(1|2|4)$ denotes a partial ranking without ties; and, $\sigma_4=(1,2|4)$ denotes a partial ranking with ties.

Given a dataset composed of $n$ rankings $\sigma_1 ,\sigma_2 ,\ldots, \sigma_n$ provided by a set of $n$ agents, the rank aggregation problem (RAP) aims to identify the consensus permutation that best represents this dataset \cite{Dwork2001}. The consensus permutation is a permutation in which its difference to the rankings of the dataset is minimal. The difference between the two rankings is usually measured by the distance. Among the distance measures available in the literature, the Kendall tau distance (or the Kendall distance) \cite{Kendall1938} is the most widely used in several real-world applications centered on the analysis of ranked data \cite{Aledo2013,Zhou2018,Alfaro2021,Rodrigo2021}.

The Kendall distance between two permutations counts the total number of pairs of labels that are assigned to different relative orders in these two rankings. Formally, given two permutations $\pi_u$ and $\pi_v$, the Kendall distance $d(\pi_u ,\pi_v)$ can be defined as follows:
\begin{equation}\label{Equ:Kendall Tau Distance}
\begin{aligned}
d(\pi_u ,\pi_v) = \left| {\{(i,j):i < j,(\pi_u(i) > \pi_u(j)~\wedge~} \right. \\
 \left. {\pi_v(i) <\pi_v(j)) \vee (\pi_u(i) < \pi_u(j)~\wedge~\pi_v(i) > \pi_v(j))\}} \right| \\
 \end{aligned}
\end{equation}

This is an intuitive and easily interpretable measure. For two arbitrary permutations, the time complexity of computing the Kendall distance $d(\pi_u ,\pi_v)$ is $O(m \log (m))$.

To calculate the distance between two arbitrary rankings $\sigma_u$ and $\sigma_v$, the extended Kendall distance $d'(\sigma_u ,\sigma_v)$ counts the total number of label pairs over which they disagree, ignoring the label pairs that are not ranked in both $\sigma_u$ and $\sigma_v$. Considering the four aforementioned rankings, $d'(\sigma_3,\sigma_4)=0$, $d'(\sigma_2,\sigma_3) = 1$, and $d'(\sigma_1,\sigma_3)=1$.

Given a set of arbitrary rankings $\{\sigma_1,\sigma_2,\ldots, \sigma_n\}$ over $m$ labels $\mathcal{L}_m = \{\lambda_1,\lambda_2,\ldots,\lambda_m\}$, RAP aims to find the permutation $\pi_o$ such that
\begin{equation}\label{Equ:Objective Function}
\pi_o \leftarrow \arg \min_{\pi \in \Omega} \frac{1}{n} \sum^n_{i=1}d'(\pi,\sigma_i)
\end{equation}
where $\Omega$ denotes the permutation space of $\{1,2,\ldots,m\}$ and $d'(\pi,\sigma_i)$ denotes the extended Kendall distance between the two rankings $\pi$ and $\sigma_i$. $\pi_o$ is the consensus permutation that minimizes the sum of the total number of pairwise disagreements with respect to the given rankings.

When all the rankings in the dataset are permutations, RAP becomes the well-established Kemeny ranking problem (KRP) \cite{Bartholdi1989,Davenport2004}. In general, RAP handles both complete and partial rankings. In both cases, the solution of RAP is a permutation (i.e., complete ranking without ties).

Solving RAP is computationally challenging because it is known to be NP-hard, when aggregating only four rankings \cite{Bartholdi1989}. As the review presented in Section \ref{Sec:Related Work} indicates, even if several solution methods have been proposed for the problem, there is still room for improvement. Certainly, the best existing methods for RAP with complete rankings are time consuming for solving large RAP instances. For RAP with partial rankings, only local search algorithms have been proposed. To the best of our knowledge, a powerful population-based memetic approach \cite{Neri2012} has not yet been studied for RAP with arbitrary rankings.

In this study, we develop an effective hybrid evolutionary ranking (HER) algorithm for solving the RAP (see Section \ref{Sec:Memetic Search for RAP}). The proposed algorithm features two  original and complementary search components: a concordant pair-based semantic crossover (CPSC) to construct meaningful offspring solutions, and an efficient late acceptance driven search (LADS) to find high-quality local optima. The main contributions of this study are summarized as follows.

From the perspective of algorithm design, the proposed CPSC crossover is the first backbone-based crossover for RAP that relies on the identification and transmission of concordant pairs (building blocks) shared by the parent solutions. By inheriting meaningful building blocks, CPSC aims to generate promising offspring solutions that serve as starting points for local optimization. The local optimization component explores different high-quality solutions around each new offspring solution owing to the combined use of a late acceptance strategy and the first efficient incremental evaluation technique introduced for RAP.

From the perspective of computational results, we present extensive experimental studies to demonstrate the high competitiveness of the proposed algorithm compared to state-of-the-art algorithms on popular benchmark instances.In addition, we show the practical usefulness of this study for an important machine learning problem related to label ranking.

The remainder of this paper is organized as follows: Section \ref{Sec:Related Work} provides a review of existing rank aggregation methods. Section \ref{Sec:Memetic Search for RAP} presents the proposed algorithm, followed by the computational results and comparisons in Section \ref{Sec:Computational Studies}. The practical usefulness of the proposed method is illustrated in label ranking in Section \ref{Sec:Applied to Label Ranking}. Experimental studies on the key issues of the proposed algorithm are presented in Section \ref{Sec:Discussion and Analysis}. Section \ref{Sec:Concluding Remarks} summarizes the study's contributions.

\section{Related Work on Rank Aggregation}
\label{Sec:Related Work}

Owing to the theoretical and practical significance of RAP, considerable effort has been devoted to the design of solution methods for this problem. These methods can be classified into two categories: exact algorithms and heuristic algorithms. Ali and Meil{\u{a}} \cite{Ali2012} performed an experimental analysis of many heuristics and exact algorithms for solving the RAP problem using data obtained from Mallows distributions following different parameterizations. Because RAP is an NP-hard problem, exact algorithms are only practical for problem instances with a limited size. To handle large and difficult instances, several heuristic algorithms have been proposed to find approximate solutions. Existing heuristic algorithms can be further divided into two categories according to the type of RAP.

\textbf{Rank aggregation for complete rankings}: The standard Borda method \cite{Borda1781} is a well-established greedy heuristic for RAP, which is intuitive and simple to compute for complete rankings. This method has the advantage of being simple and fast, but the obtained solutions may be far from the true optima. Aledo et al \cite{Aledo2013} used the genetic algorithm (GA) to solve the RAP problem with complete rankings (i.e., KRP). Even if this algorithm only relies on standard permutation crossovers (position-based crossover, order crossover, order-based crossover) and mutations (insertion, displacement, and inversion), it obtained significantly better results than the most representative algorithms studied in \cite{Ali2012}. Aledo et al \cite{Aledo2018} further applied $(1+\lambda)$ evolution strategies (ES) to solve the optimal bucket order problem (OBOP), whose objective is to obtain a complete consensus ranking (ties are allowed) from a matrix of preferences. They experimentally evaluated several configurations of the designed ES algorithm. It is worth noting that these algorithms were developed for RAP with complete rankings only.

\textbf{Rank aggregation for partial rankings}: To address RAP in the general setting, Aledo et al \cite{Aledo2016} proposed an improved Borda method for RAP containing any kind of rankings and outperformed the standard Borda method. In addition, N{\'a}poles et al \cite{Napoles2017} applied ant colony optimization to solve an extension of KRP, that is, the weighted KRP for partial rankings. D’Ambrosio et al \cite{DAmbrosio2017} proposed a differential evolution algorithm for consensus-ranking detection within Kemeny's axiomatic framework. Recently, Aledo et al \cite{Aledo2019} performed a comparative study of four local search-based algorithms: hill climbing (HC), iterated local search (ILS), variable neighborhood search (VNS) and greedy randomized adaptive search procedure (GRASP). Both the interchange and insert neighborhood are used in these local search algorithms. The comparative results showed that GRASP can achieve the best tradeoff between accuracy and efficiency when the algorithms are allowed to perform a large number of fitness evaluations.

\section{Hybrid Evolutionary Search for Rank Aggregation Problem}
\label{Sec:Memetic Search for RAP}

In this section, we present the first hybrid evolutionary ranking (HER) algorithm for the rank aggregation problem with complete rankings. We begin with the solution representation and evaluation and then introduce the main components of the proposed algorithm. In Section \ref{SubSec:Results on RAP With Partial Rankings}, we explain how the algorithm can be easily adapted to the case of partial rankings by simply replacing the Kendall distance with the extended Kendall distance.

\subsection{Solution Representation and Evaluation}
\label{SubSec:Solution Representation and Evaluation}

Let $\mathcal{D} = \{\pi_1 ,\pi_2 ,\ldots, \pi_n\}$ be a given dataset, and a feasible candidate solution for the problem is a permutation $\pi$ of the set $\{1,2,\ldots,m\}$. The search space $\Omega$ is composed of all possible permutations of size $n$. For a given candidate solution $\pi$ in $\Omega$, the objective function value (fitness) is calculated as follows:
\begin{equation}
\label{function evaluation}
    f(\pi) = \frac{1}{n} \sum_{k=1}^{n} d(\pi,\pi_k)
\end{equation}
where $d(\pi,\pi_k)$ denotes the Kendall distance between $\pi$ and $\pi_k$. Because calculating the Kendall distance $d(\pi,\pi_k)$ requires $O(m \log (m))$ time, the evaluation of a candidate solution requires $O(n \cdot m\log (m))$ time. The purpose of the HER algorithm is to find a permutation $\pi^* \in \Omega$ with the smallest objective function value $f(\pi^*)$.

\subsection{General Framework}
\label{SubSec:General Framework}

The HER algorithm follows the memetic algorithm framework in discrete optimization \cite{Hao2012} and combines a population-based approach with local optimization. As shown in Algorithm \ref{Alg:Pseudo-Code HES Approach}, the HER is composed of four main components: a population initialization procedure, a concordance pairs-based semantic crossover (CPSC), a late acceptance driven search (LADS), and a population updating strategy. The algorithm starts with a population of high-quality solutions. At each subsequent generation, a promising offspring solution is first generated by CPSC, then improved by the LADS procedure, and finally considered for acceptance by the population updating strategy. The process is repeated until a stopping condition (i.e., the time limit $t_{max}$ or allowable maximum number of generations without improvement $MaxGens$) is satisfied. We present each key procedure in the following sections.

\begin{algorithm}[!ht]
\small
\caption{Hybrid Evolutionary Search for RAP}
\label{Alg:Pseudo-Code HES Approach}
\begin{algorithmic}[1]
    \REQUIRE{Problem instance $I$, population size $T$, and maximum number of generations without improvement $MaxGens$}
    \ENSURE{The best-found solution $\pi^*$}
 	//Build an initial population of high-quality solutions \\
    \STATE $\textit{P} \leftarrow \texttt{PopulationInitialization}(T)$;\\
    //Record the best solution\\
    \STATE $\pi^{*} \leftarrow \arg \min_{\pi_i \in P} f(\pi_i)$;\\
    \STATE $idle\_gens \leftarrow 0$;
	\WHILE{Stopping condition is not met}
		\STATE//Construct an offspring solution; \\
        \STATE $\pi' \leftarrow \texttt{ConcordantPairBasedSemanticCrossover}(P)$;\\
		//Improve it through local optimization; \\
        \STATE $\pi' \leftarrow \texttt{LateAcceptanceDrivenSearch}(\pi', \textit{MaxIters})$;\\
        //Update the best solution\\
		\IF{$f(\pi') < f(\pi^*)$}
			\STATE $\pi^{*} \leftarrow \pi'$;\\
            \STATE $idle\_gens \leftarrow 0$;\\
        \ELSE
            \STATE $idle\_gens \leftarrow idle\_gens+1$;\\
        \ENDIF
        \IF{$idle\_gens > MaxGens$}
            \STATE \textbf{break};\\
        \ENDIF
        \\//Update the population;\\
        \STATE $P \leftarrow \texttt{PopulationUpdating}(P,\pi')$;\\
	\ENDWHILE
    \RETURN The best found solution $\pi^*$
\end{algorithmic}
\end{algorithm}

\subsection{Population Initialization}
\label{SubSec:Population Initialization}

HER starts its search with a population of high-quality solutions, where each solution is obtained in two steps. First, an initial solution is obtained using an improved Borda procedure. Then, the initial solution is further improved by LADS (see Section \ref{SubSec:Late Acceptance Driven Search}) before being added to the population.

The Borda procedure uses a well-established voting rule in social choice theory. We assume that the preferences of $n$ voters are expressed in terms of rankings $\pi_1 ,\pi_2 ,\ldots, \pi_n$ over $m$ alternatives. For each ranking $\pi_i$, the best alternative receives $m-1$ points, and the second best receives $m-2$ points. The total score of an alternative is the sum of the points that it has received from all $n$ voters. Finally, a representative ranking is obtained based on the scores of the alternatives. In other words, all alternatives are sorted in decreasing order of their scores, and the ties are broken at random. The Borda method is simple and terminates in $O(nm)$ time. To introduce randomness into solutions, which is helpful for effective exploration of the search space, we adopt a randomized Borda method that aggregates only $(1-\beta) \cdot n$ ($\beta \in(0,0.5)$ is a randomized factor) rankings randomly selected from $n$ rankings.

\subsection{Concordant Pairs-based Semantic Crossover}
\label{SubSec:Concordant Pairs-Based Semantic Crossover}

As a driving force of hybrid evolutionary algorithms, a meaningful crossover operator should be able to generate promising offspring solutions that not only inherit the good properties of the parents but also introduce new useful characteristics \cite{Hao2012}. The concept of backbone has been widely used to define the good properties of parents. A variety of backbone-based crossovers have been proposed for subset selection problems, such as the maximum diversity problem \cite{Wu2013,Zhou2017}, Steiner tree problem \cite{Fu2015}, critical node problem \cite{Zhou2019,Zhou2021}, grouping problems such as graph coloring \cite{Galinier1999}, and generalized quadratic multiple knapsack problem \cite{Chen2016}. For the RAP problem whose solutions are permutations, we propose the first backbone-based crossover, which relies on the identification and transmission of concordant pairs (building blocks) shared by the parent solutions. By inheriting meaningful building blocks, crossover favors the generation of promising offspring solutions.

Given a ranking $\pi$ of $m$ labels, it can be equivalently transformed into a set of $m(m-1)/2$ pairwise preferences. For example, from $\pi = (4|2|1|3)$, we obtain a set of $4 \times (4-1)/2$ pairwise preferences $\{\lambda_4 \prec \lambda_2,\lambda_4 \prec \lambda_1, \lambda_4 \prec \lambda_3, \lambda_2 \prec \lambda_1, \lambda_2 \prec \lambda_3, \lambda_1 \prec \lambda_3\}$ where $\prec$ is the preference relation. Therefore, for any two or more rankings, their backbone can be defined as a set of concordant pairs (see Definition \ref{Def:Concordant Pairs}).

\begin{def1}{(\textbf{Concordant pairs}).}\label{Def:Concordant Pairs}
Given two rankings $\pi_u$ and $\pi_v$ of $m$ labels, a pair of labels $(\lambda_i,\lambda_j)$ is a concordant pair if labels $\lambda_i$ and $\lambda_j$ share the same preference relation $\lambda_i \prec \lambda_j$ or $\lambda_j \prec \lambda_i$ in the parent rankings.
\end{def1}

Given two parent rankings $\pi_u$ and $\pi_v$ randomly selected from the population $\mathcal{P}$, the CPSC operator builds an offspring solution $\pi_o$ in four steps.
\begin{itemize}
    \item \textbf{Step 1:} decompose each parent ranking $\pi_k, k \in \{u,v\}$ into a set of $|\pi_k|\cdot(|\pi_k|-1)/2$ pairwise preference relations $\mathcal{R}_{k}$;
    \item \textbf{Step 2:} identify all concordant pairs (i.e., common preference relations between parent rankings), that is, $\mathcal{R}_o \leftarrow \mathcal{R}_u \bigcap \mathcal{R}_v$;
    \item \textbf{Step 3:} combine the concordant pairs into a partial ranking according to a voting strategy, that is, $\pi_o \xleftarrow{\emph{a voting strategy}} \mathcal{R}_o$;
    \item \textbf{Step 4:} complete $\pi_o$ to form a feasible solution (i.e., a permutation) by determining all unknown preference relations in a random manner.
\end{itemize}

After identifying all concordant preference relations $\mathcal{R}_o$ between the parent rankings, the next question is how to derive an associated ranking based on $\mathcal{R}_o$. This question is nontrivial, because a relation $\mathcal{R}_o$ does not always suggest a unique ranking. The CPSC first maps the identified $\mathcal{R}_o$ into a partial ranking according to a voting strategy (breaking ties randomly), and then repairs it to a feasible ranking.

\begin{figure}[!ht]
\centering
\includegraphics[width=1.0\columnwidth]{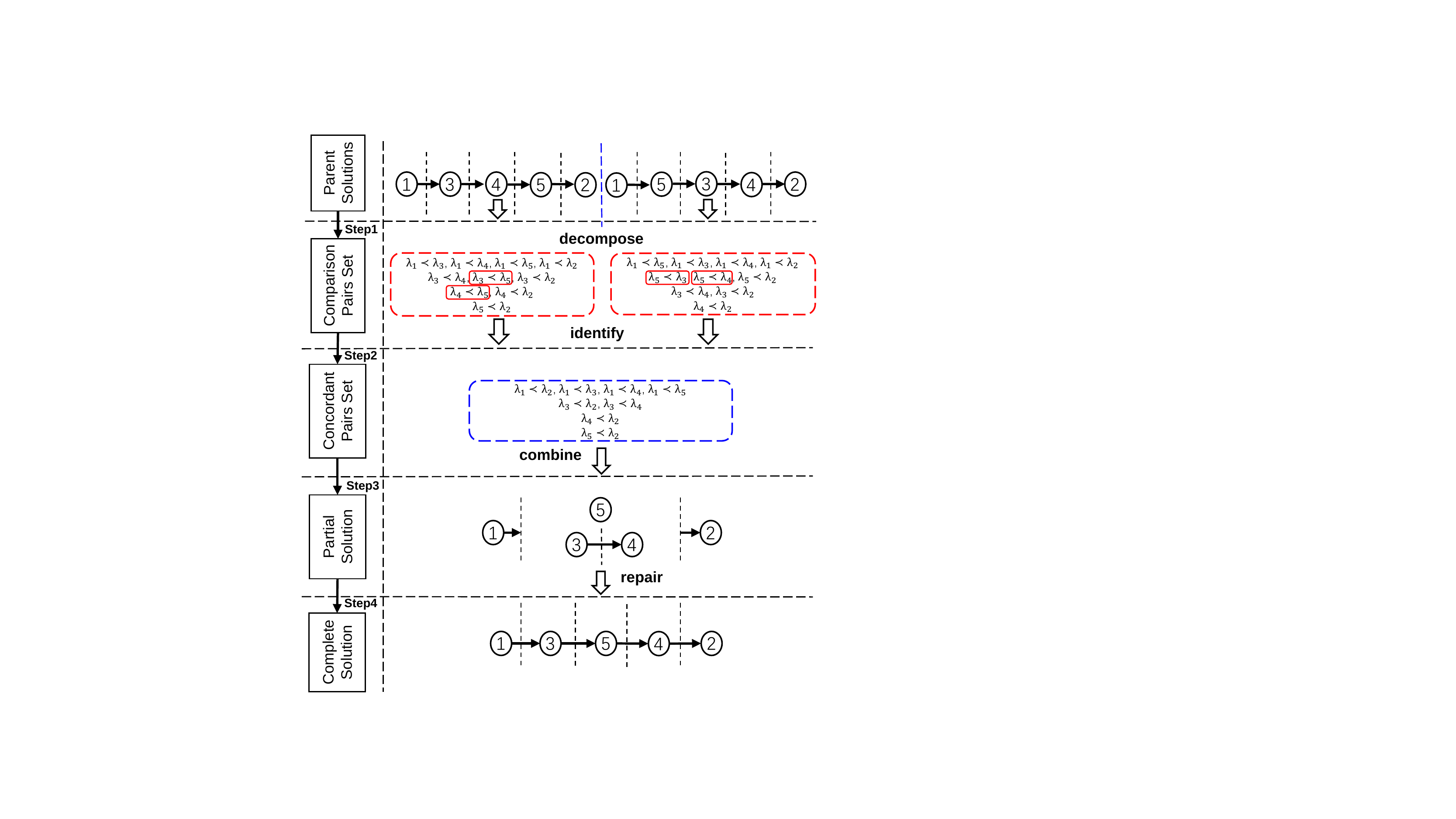}
\caption{Illustration of the Concordant Pairs-based Semantic Crossover}
\label{Fig:Schematic Illustration of Constructing Offspring Solutions by CPSC}
\end{figure}

Figure~\ref{Fig:Schematic Illustration of Constructing Offspring Solutions by CPSC} shows an illustrative example of the CPSC crossover with two parent solutions: $\pi_1 = (1|3|4|5|2)$ and $\pi_2 = (1|5|3|4|2)$. \textbf{Step 1} decomposes the parent solutions into two sets of pairwise preference relation pairs: $\mathcal{R}_1=\{\lambda_1 \prec \lambda_3,\lambda_1 \prec \lambda_4, \lambda_1 \prec \lambda_5, \lambda_1 \prec \lambda_2, \lambda_3 \prec \lambda_4, \lambda_3 \prec \lambda_5, \lambda_3 \prec \lambda_2, \lambda_4 \prec \lambda_5, \lambda_4 \prec \lambda_2, \lambda_5 \prec \lambda_2\}$ and $\mathcal{R}_2=\{\lambda_1 \prec \lambda_5,\lambda_1 \prec \lambda_3, \lambda_1 \prec \lambda_4, \lambda_1 \prec \lambda_2, \lambda_5 \prec \lambda_3, \lambda_5 \prec \lambda_4, \lambda_5 \prec \lambda_2, \lambda_3 \prec \lambda_4, \lambda_3 \prec \lambda_2, \lambda_4 \prec \lambda_2\}$. \textbf{Step 2} identifies all concordant pairs $\mathcal{R}_0=\{\lambda_1 \prec \lambda_2,\lambda_1 \prec \lambda_3, \lambda_1 \prec \lambda_4, \lambda_1 \prec \lambda_5, \lambda_3 \prec \lambda_2, \lambda_3 \prec \lambda_4, \lambda_4 \prec \lambda_2, \lambda_5 \prec \lambda_2\}$ (i.e., common preference relation pairs) between $\mathcal{R}_1$ and $\mathcal{R}_2$, which form the backbone of the parent solutions. \textbf{Step 3} combines the concordant pairs $\mathcal{R}_0$ into a partial ranking $\pi_0 = (1|5,3|4|,|2)$. \textbf{Step 4} repairs $\pi_0$ to obtain a feasible solution (i.e., a permutation) $\pi_0 = (1|3|5|4|2)$. Specifically, we determine the unknown preference relation between 5 and $(3|4)$ in a random manner by following all existing preference relation pairs $(5|3|4)$, $(3|5|4)$, and $(3|4|5)$ (in our example, $(3|5|4)$ is considered).

\subsection{Late Acceptance Driven Search}
\label{SubSec:Late Acceptance Driven Search}

In addition to the CPSC operator, HER relies on a highly effective local optimization procedure, which employs a late acceptance strategy, reinforced by a fast incremental evaluation mechanism. We describe the components of this late acceptance driven search procedure as follows.

\subsubsection{Neighborhood Structure.}
\label{SubSubSec:Neighborhood Structure}

The neighborhood of a local search is typically defined by a move operator, which transforms the current solution to generate a neighboring solution \cite{Samorani2012,Wu2021}. For RAP, to generate a neighboring ranking $\pi'$ from a given ranking $\pi$, the LADS procedure uses the basic swap operator (\texttt{SWAP}), which exchanges two different positions of $i$ and $j$ of $\pi$. This move operation is denoted by $\pi' \leftarrow \pi \oplus \texttt{SWAP}(i,j)$.

Given a permutation $\pi$, the set of neighboring solutions $\pi'$ of the swap neighborhood $N$ is then represented by
\begin{equation*}
    N(\pi) = \{\pi' \leftarrow \pi \oplus \texttt{SWAP}(i,j), \forall i \neq j \in \{1,\ldots,n\}\}
 \end{equation*}

For example, $(1|4|3|2|5)$ is a neighboring ranking of ranking $(1|2|3|4|5)$ by swapping the positions of labels 4 and 2. Clearly, the size of the swap neighborhood $N(\pi)$ is $n(n-1)/2$.

\subsubsection{Incremental Evaluation of Neighboring Solutions.}

Evaluating a neighboring solution according to Equation (\ref{function evaluation}) requires $O(n \cdot m\log(m))$ time complexity, which is extremely time-consuming. It is worth noting that all existing algorithms for RAP suffer from the high computational complexity of calculating the Kendall distance. Typically, the objective function value of a candidate neighboring solution must be computed from scratch, which considerably slows down the search process, particularly for large instances.

To overcome this problem, we propose the first incremental evaluation technique to speed up the computation of the objective function for the RAP. Given the Kendall distance $d(\pi,\pi_k)$ between a candidate ranking $\pi$ and a given ranking $\pi_k$, we assume $\pi'$ be a neighboring solution of $\pi$ by performing a swap operation between two different positions $i$ and $j$ of $\pi$, i.e., $\pi' \leftarrow \pi \oplus \texttt{SWAP}(i,j), i \neq j \in \{1,\ldots,n\}$. Then, the Kendall distance $d(\pi',\pi_k)$ between $\pi'$ and $\pi_k$ can be incrementally calculated as follows:
\begin{equation}
    d(\pi',\pi_k) = d(\pi,\pi_k)+\Delta d(\pi,\pi',\pi_k,i,j)
\end{equation}
where the calculation of $\Delta d(\pi,\pi',\pi_k,i,j)$ described in Algorithm \ref{Alg:Incremental Evaluation Procedure}.

Using this incremental evaluation technique, we can compute the objective function value of a neighboring solution more efficiently as follows:
\begin{equation}
f(\pi') = f(\pi)+\frac{1}{n}\sum_{i=1}^{n}\Delta d(\pi,\pi',\pi_k,i,j)
\end{equation}
This reduces the complexity from $O(n \cdot m\log(m))$ to $O(n \cdot m)$.

\begin{algorithm}[!ht]
\scriptsize
\caption{Incremental Evaluation Technique for Calculating $\Delta d(\pi,\pi',\pi_k,i,j)$}
\label{Alg:Incremental Evaluation Procedure}
\begin{algorithmic}[1]
 	\REQUIRE{A given ranking $\pi_k$, a candidate ranking $\pi$ and its neighboring ranking $\pi'$ that is obtained by performing $\texttt{SWAP}(i,j)$ operation on $\pi$}
 	\ENSURE{The incremental distance $\Delta d(\pi,\pi',\pi_k,i,j)$}
    \STATE $\textit{count} \leftarrow 0$;\\
    \IF{$\pi_k(i) > \pi_k(j)$}
        \STATE $\texttt{SWAP}(i,j)$;\\
    \ENDIF
    \IF{$\pi(i) > \pi(j)$}
        \STATE $\textit{count} \leftarrow \textit{count}+1$;\\
    \ELSE
        \STATE $\textit{count} \leftarrow \textit{count}-1$;\\
    \ENDIF
    \FOR{$\forall \textit{temp} \in (\pi_k(i),\pi_k(j))$}
        \STATE $v \leftarrow \arg \pi_k(v)= \textit{temp}$;\\
        \IF{$(\pi(i) > \pi(v)$~and~$\pi(j) < \pi(v))$~or~$(\pi(i) < \pi(v)$~and~$\pi(j) > \pi(v))$}
            \IF{$\pi(i) > \pi(v)$}
                \STATE $\textit{count} \leftarrow \textit{count}-1$;\\
            \ELSE
                \STATE $\textit{count} \leftarrow \textit{count} + 1$;\\
            \ENDIF
            \IF{$\pi(v) > \pi(j)$}
                \STATE $\textit{count} \leftarrow \textit{count} - 1$;\\
            \ELSE
                \STATE $\textit{count} \leftarrow \textit{count} + 1$;\\
            \ENDIF
        \ENDIF
    \ENDFOR
    \STATE $\Delta d(\pi,\pi',\pi_k,i,j) \leftarrow \textit{count}$;\\
    \RETURN The incremental distance $\Delta d(\pi,\pi',\pi_k,i,j)$
\end{algorithmic}
\end{algorithm}

\subsubsection{Late Acceptance Strategy.}
\label{SubSubSec:Late Acceptance Strategy}

The late acceptance strategy \cite{Burke2008} extends the well-established HC algorithm. At each step of the HC, a candidate solution is always compared with the current solution. The late acceptance strategy delays the comparison, where a new candidate solution is compared with one of some pre-encountered solutions. Based on the late acceptance strategy, several effective HC-based algorithms have been proposed \cite{Burke2017,Namazi2018}. However, as indicated by Namazi et al \cite{Namazi2018} and Zhou et al \cite{Zhou2021}, they are generally time-consuming to achieve a good result. To speed up the search, we propose a LADS, which effectively integrates the above incremental evaluation technique to evaluate a candidate solution in an incremental manner. LADS stores the solution costs of a predefined number of previous iterations in a cost list of length $L_h$. LADS accepts a non-improving candidate solution if it has a higher cost than the previous cost stored in the cost list. Algorithm \ref{Algorithm:LADS} presents the LADS procedure.

\begin{algorithm}[!ht]
\small
\scriptsize
\caption{Late Acceptance Driven Search}
\label{Algorithm:LADS}
\begin{algorithmic}[1]
 	\REQUIRE{Initial solution $\pi$ and allowable maximum number of iterations without improvement $\textit{MaxIters}$}
 	\ENSURE{The best solution $\pi^*$ found}
    \STATE Initialize the history length $L_h$;\\
    \STATE $\pi^* \leftarrow \pi$;\\
    \FOR{$\forall i \in \{0,\ldots,L_h-1\}$}
        \STATE $f_i \leftarrow f(\pi)$;
    \ENDFOR
    \STATE $f_{max} \leftarrow f(\pi)$, $\textit{count} \leftarrow L_h$;\\
    \STATE Initialize $\textit{iters} \leftarrow 0$, $\textit{idle\_iters} \leftarrow 0$;\\
	\WHILE{$\textit{idle\_iters} < \textit{MaxIters}$}
        \STATE $f_{prev} \leftarrow f(\pi)$;\\
        /*Generate a neighboring solution*/\\
        \STATE $\pi' \leftarrow \pi \oplus \texttt{SWAP}(x,y)$;\\
        /*Evaluate it using the incremental evaluation technique*/\\
        \STATE $f(\pi') \leftarrow f(\pi)+\sum_{k=1}^{n}\Delta d(\pi,\pi',\pi_k,x,y)$;\\
        /*Calculate the virtual beginning*/\\
        \STATE $v \leftarrow \textit{iters} \mod L_h$;\\
        \IF{$f(\pi') = f(\pi)$ or $f(\pi') < f_{max}$}
            \STATE $\pi \leftarrow \pi'$, $f(\pi) \leftarrow f(\pi')$;\\
            /*Update the best solution*/\\
            \IF{$f(\pi) < f(\pi^*)$}
                \STATE $\pi^* \leftarrow \pi$, $f(\pi^*) \leftarrow f(\pi)$;\\
                \STATE $\textit{idle\_iters} \leftarrow 0$;
            \ELSE
                \STATE $\textit{idle\_iters} \leftarrow \textit{idle\_iters} + 1$;
            \ENDIF
        \ENDIF
        \\/*Update the cost list*/\\
        \IF{$f(\pi) > f_v$}
            \STATE $f_v \leftarrow f(\pi)$;
        \ELSE
            \IF{$f(\pi) < f_v$ and $f(\pi) < f_{prev}$}
                \IF{$f_v = f_{max}$}
                    \STATE $\textit{count} \leftarrow \textit{count} - 1$;
                \ENDIF
                \STATE $f_v \leftarrow f(\pi)$;\\
                \IF{$\textit{count} = 0$}
                    \STATE compute $f_{max}, \textit{count}$;\\
                \ENDIF
            \ENDIF
        \ENDIF
        \STATE $\textit{iters} \leftarrow \textit{iters} + 1$;
	\ENDWHILE
    \RETURN The best solution found $\pi^*$
\end{algorithmic}
\end{algorithm}

\subsection{Population Updating Strategy}
\label{SubSec:Population Updating Strategy}

Diversity is a property of a group of individuals that indicates how much these individuals are identical. A suitable population updating strategy is necessary to maintain population diversity during the search, thus preventing the algorithm from premature convergence and stagnation \cite{Neri2012}. Diversity is often used to determine whether the offspring solution should be inserted into the population or discarded. In this study, we adopt a simple strategy that always replaces the worst individual if the offspring has a better solution quality and is different from any existing individual in the population.

\subsection{Computational Complexity of HER}
\label{SubSec:Computational Complexity of HER}

To analyze the computational complexity of the proposed HER algorithm, we consider the main procedures in one generation in the main loop of Algorithm \ref{Alg:Pseudo-Code HES Approach}. At each generation, the HER executes three procedures: CPSC, LADS and population updating. The CPSC crossover can be performed in $O(m^2+m\log(m)+m)$. The time complexity of LADS is $(NbrIters \cdot (nm+L_h))$, where $NbrIters$ denotes the total number of iterations executed in LADS and $L_h$ denotes the history length. The computational complexity for population updating is $O(T(m^2+T))$, where $T$ is the population size. To summarize, the total computational complexity of the proposed HER for one generation is $O(m^2+nm \cdot NbrIters)$.

\section{Computational Studies}
\label{Sec:Computational Studies}

In this section, we present a computational assessment of the HER algorithm and its LADS procedure. We first describe the benchmark instances and the experimental settings. Then, we present the computational results obtained on the benchmark instances and compare them with the state-of-the-art algorithms.

\subsection{Benchmark Instances and Experimental Settings}
\label{SubSec:Benchmark Instances and Experimental Settings}

Our studies are conducted on 400 widely used benchmark instances\footnote{They are publicly available at \url{http://simd.albacete.org/rankings/}}. They were sampled from the Mallows distribution. To define a standard Mallows distribution, three parameters are required: the center permutation $\pi_0$, the spread parameter $\theta$, and the length of the permutation $m$. In addition, the number of permutations to be sampled $n$ is also needed to define a practical instance. For this category of instances, $\pi_0$ is always set to the identity permutation $\pi_0 = (1,2,\ldots,m)$, $\theta \in \{0.001, 0.01, 0.1,0.2\}$, $m \in \{50,100,150,200,250\}$, and $n = 100$. For each of the 20 combinations of $\theta$ and $m$, 20 instances with $n = 100$ permutations were generated. As indicated by Aledo et al \cite{Aledo2013}, the most complex instances are those with a small $\theta$ and a large permutation size $m$.

Our algorithms\footnote{Our programs and results will be made available at \url{https://github.com/YangmingZhou/RankAggregationProblem}}  were programmed in C++ and compiled using GNU gcc 4.1.2 with the `-O3' option on an Intel E5-2670 with 2.5GHz and 2GB RAM under the Linux OS. The detailed parameter settings of our algorithms are listed in Tables \ref{Tab:Parameter Settings}. Following \cite{Aledo2013}, we set $MaxGens$ to 60 as the stopping condition. The population size $T$ is set to $20$, as suggested by L{\"u} and Hao \cite{Lu2010} and Zhou et al \cite{Zhou2021}. Our preliminary analysis indicated that $\beta$, $MaxIters$ and $L_h$ are sensitive parameters, whereas $T$ is not. We used the general practice in heuristic algorithm design to tune $\beta$, $MaxIters$ and $L_h$ by experimentally determining them on some representative instances. As an example, we present a detailed experimental analysis of the parameter $L_h$ in Section \ref{SubSec:Effect of the History Length on LADS}.

\begin{table*}[!htp]
\centering
\small
\caption{Parameter Settings of Our Algorithms}
\label{Tab:Parameter Settings}
\begin{threeparttable}
\setlength{\tabcolsep}{3.0mm}{
\begin{tabular}{llrc}
\toprule[0.75pt]
Parameter & Description & Value & Section \\
\midrule[0.5pt]
$MaxGens$ & Maximum number of generations without improvement & 60 & Section \ref{SubSec:General Framework} \\
$T$ & Population size & 20 & Section \ref{SubSec:Population Initialization} \\
$\beta$ & Randomized factor & 0.2 & Section \ref{SubSec:Population Initialization} \\
$MaxIters$ & Maximum number of iterations without improvement & 5000 & Section \ref{SubSec:Late Acceptance Driven Search} \\
$L_h$ &Length of history costs & 5 & Section \ref{SubSec:Late Acceptance Driven Search} \\
\bottomrule[0.75pt]
\end{tabular}}
\end{threeparttable}
\end{table*}

\subsection{Comparisons with State-of-the-Art Algorithms}
\label{SubSec:Comparison of HES with State-of-the-Art Algorithms}

This section compares our HER algorithm and its LADS procedure with the following five state-of-the-art algorithms.
\begin{enumerate}
    \item \textbf{Borda} is a well-established greedy heuristic algorithm for RAP. It is simple and fast, and can perform rank aggregation in linear time $O(nm)$ \cite{Borda1781}.
    \item \textbf{CSS} is a graph-based approximate algorithm that implements a greedy version of the method introduced by Cohen et al \cite{Cohen1999}.
    \item \textbf{DK} is an exact solver proposed by Davenport and Kalagnanam \cite{Davenport2004}, and is enhanced with improved heuristics.
    \item \textbf{Branch and bound (B\&B)} relies on the well-established A* algorithm combined with admissible heuristics \cite{Ali2012}. As indicated by Aledo et al \cite{Aledo2013}, the exact version of B\&B runs out of space in most cases. Therefore, an approximate version was used.
    \item \textbf{The genetic algorithm (GA)} is a population-based algorithm for estimating the consensus permutation of rank aggregation problems, which achieves state-of-the-art results on instances from the Mallows model Aledo et al \cite{Aledo2013}.
\end{enumerate}

Borda, CSS, and DK are greedy algorithms. They are considerably faster than B\&B and GA, but often produce poor results. GA is far slower than B\&B because the large number of fitness evaluations required during the evolutionary search. As shown in \cite{Aledo2013}, the CPU time ratios (i.e., $\frac{t_{GA}}{t_{B\&B}}$) between the GA and B\&B are 9.6, 15.6, 219.4, and 639.9 on four extreme instances (i.e., four combinations between $\theta \in \{0.001,0.2\}$ and $m \in \{50,250\}$). With the condition that GA stops after 60 generations without improving the best solution, GA achieved state-of-the-art results on the benchmark instances \cite{Aledo2013}. Following the literature \cite{Aledo2013}, we solve each instance once and terminate the HER algorithm after 60 generations without improving the best solution or the execution time reaches the time limit $\hat{t}=2$ hours. Our stopping condition is much stricter than that of the GA. We then obtained the best result ($f_{best}$), the average result ($f_{avg}$) and the average time ($t_{avg}$) over each group of 20 instances. We also use the Wilcoxon signed-rank test for the comparison of two algorithms, as recommended in \cite{Demvsar2006}. The results of our algorithms (i.e., LADS and HER) and the reference algorithms are summarized in Table \ref{Tab:Comparison of Our Algorithms with State-of-the-Art Algorithms on Dataset With Complete Rankings}.

\begin{table*}[!hbtp]
\centering
\caption{Comparison of Our Algorithms with the Reference Algorithms on Each Combination of $\theta$ and $m$}
\label{Tab:Comparison of Our Algorithms with State-of-the-Art Algorithms on Dataset With Complete Rankings}
\begin{threeparttable}
\resizebox{\textwidth}{!}{
\begin{tabular}{lc|rrrrr|rrr|rrr}
\toprule[0.75pt]
\multicolumn{2}{c|}{Instance}  & Borda & CSS &DK & B\&B & GA & \multicolumn{3}{c|}{LADS} & \multicolumn{3}{c}{HER}\\
\midrule[0.5pt]
$\theta$ & $m$ & $f_{best}$ & $f_{best}$ & $f_{best}$ & $f_{best}$ & $f_{best}$ & $f_{best}$ & $f_{avg}$ & $t_{avg}$ & $f_{best}$ & $f_{avg}$ & $t_{avg}$\\
\midrule[0.5pt]
0.200 &050&187.837&188.342&187.816&187.815&187.815&\textbf{183.140} &187.914 &2.633 &\textbf{183.140} &\textbf{187.913} &3.662 \\
0.100 &050&320.194&320.883&320.128&320.104&320.104&\textbf{311.950} &320.304 &2.446 &\textbf{311.950} &\textbf{320.296} &4.767 \\
0.010 &050&559.915&560.720&559.582&558.928&558.769&551.030 &559.755 &2.035 &\textbf{550.970} &\textbf{559.607} &434.461 \\
0.001 &050&569.701&570.499&569.546&568.662&568.469&561.760 &569.906 &1.973 &\textbf{561.480} &\textbf{569.718} &405.243 \\
0.200 &100&412.571&413.201&412.554&412.554&412.154&\textbf{405.140} &411.888 &18.735 &\textbf{405.140} &\textbf{411.884} &31.638 \\
0.100 &100&788.279&790.126&788.058&788.102&788.026&776.060 &787.779 &16.392 &\textbf{776.020} &\textbf{787.742} &89.533 \\
0.010 &100&2155.301&2157.450&2154.294&2152.986&2152.247&2126.610 &2152.856 &15.056 &\textbf{2126.390} &\textbf{2152.623} &3166.568 \\
0.001 &100&2308.277&2310.266&2308.038&2304.983&2303.231&2290.570 &2305.656 &15.254 &\textbf{2290.190} &\textbf{2305.296} &4277.235 \\
0.200 &150&637.245&638.903&637.177&637.177&637.176&629.430 &636.956 &54.732 &\textbf{629.410} &\textbf{636.948} &308.377 \\
0.100 &150&1260.964&1264.171&1260.645&1260.610&1260.583&1245.810 &1260.194 &45.819 &\textbf{1245.730} &\textbf{1260.146} &619.723 \\
0.010 &150&4595.137&4599.431&4593.498&4590.917&4589.672&4533.030 &4585.319 &118.519 &\textbf{4532.710} &\textbf{4585.154} &4610.010 \\
0.001 &150&5206.998&5210.015&5208.340&5201.233&5196.731&5143.660 &5189.971 &116.381 &\textbf{5143.140} &\textbf{5189.620} &4126.092 \\
0.200 &200&862.707&865.154&862.650&862.685&862.648&851.330 &862.704 &120.458 &\textbf{851.290} &\textbf{862.693} &1672.241 \\
0.100 &200&1734.810&1739.429&1734.336&1734.395&1734.303&1711.400 &1734.219 &103.823 &\textbf{1711.320} &\textbf{1734.181} &2263.225 \\
0.010 &200&7699.995&7706.323&7697.136&7694.639&7692.271&\textbf{7624.120} &\textbf{7697.132} &280.606 &7624.140 &7697.138 &3000.011 \\
0.001 &200&9250.210&9253.655&9256.021&9241.557&9232.840&\textbf{9179.980} &9237.531 &211.585 &9180.140 &\textbf{9237.178} &4051.452 \\
0.200 &250&1087.719&1090.796&1087.623&1087.654&1087.622&\textbf{1075.790} &1087.697 &205.948 &\textbf{1075.790} &\textbf{1087.684} &2917.056 \\
0.100 &250&2207.223&2213.130&2206.631&2206.665&2206.564&2186.480 &2206.800 &187.542 &\textbf{2186.260} &\textbf{2206.732} &3751.534 \\
0.010 &250&11311.189&11319.547&11307.085&11303.063&11300.249&\textbf{11180.290} &\textbf{11299.402} &408.436 &11180.350 &11299.597 &4170.573 \\
0.001 &250&14448.840&14451.179&14453.746&14435.551&14422.276&\textbf{14333.660} &14430.421 &422.379 &14333.840 &\textbf{14430.309} &4253.846 \\
\midrule[0.5pt]
\#Wins &&20&20&20&20&20&12&18&$-$&$-$&$-$&$-$\\
\#Ties &&0 & 0&0 &0 &0 &4 &0 &$-$&$-$&$-$&$-$\\
\#Loses&&0 & 0&0 &0 &0 &4 &2 &$-$&$-$&$-$&$-$\\
\midrule[0.5pt]
p-value&& \textbf{8.858e-5} & \textbf{8.858e-5} & \textbf{8.858e-5} & \textbf{8.858e-5} & \textbf{8.858e-5} & \textbf{2.970e-2} & \textbf{1.300e-3} &$-$&$-$&$-$&$-$\\
\bottomrule[0.75pt]
\end{tabular}}
\footnotesize
\begin{tablenotes}
    \item [$\star$] The results of each combination of $\theta$ and $m$ are averaged over 20 instances.
\end{tablenotes}
\end{threeparttable}
\end{table*}

In Table \ref{Tab:Comparison of Our Algorithms with State-of-the-Art Algorithms on Dataset With Complete Rankings}, columns 1 and 2, describe $\theta$ and $m$ values for each combination, respectively. Columns 3-7 list the best results ($f_{best}$) of the reference algorithms Borda, CSS, DK, B\&B, and GA. Because their source codes are not available, we list their results provided in \cite{Aledo2013}. Columns 8-10 list the results of the LADS procedure, including the best result ($f_{best}$) over 20 instances, the average result ($f_{avg}$), and the average time in seconds ($t_{avg}$) needed to achieve the best result for each instance. Correspondingly, columns 11-12 list the results of the HER. The best values for each performance indicator are highlighted in bold. In addition, we provide the number of combinations on which HER obtains a better (\#Wins), equal (\#Ties), and worse (\#Loses) results in terms of each indicator compared to the corresponding algorithms. At the end of Table \ref{Tab:Comparison of Our Algorithms with State-of-the-Art Algorithms on Dataset With Complete Rankings}, we also show the p-values of the Wilcoxon signed-rank test.

Table \ref{Tab:Comparison of Our Algorithms with State-of-the-Art Algorithms on Dataset With Complete Rankings} indicates that our algorithms (LADS and HER) demonstrate excellent performances for all 20 combinations of $\theta$ and $m$. At a significance level of 0.05, both LADS and HER significantly outperform the reference algorithms (i.e., Borda, CSS, DK, B\&B, GA) in terms of $f_{best}$. Compared to LADS, HER shows significantly better performances in terms of both $f_{best}$ and $f_{avg}$ at a significance level of 0.05. We also observe that LADS converges to a local optimum in approximately 400s, whereas HER has a better long-term search ability by improving its results until about 4000s. These observations confirm the competitiveness of the proposed algorithms compared to the reference algorithms.

\subsection{Results on RAP With Partial Rankings}
\label{SubSec:Results on RAP With Partial Rankings}

To extend the HER algorithm to solve the RAP with partial rankings, the objective function must be updated. Given a dataset with partial rankings $\mathcal{D}=\{\sigma_1,\sigma_2,\ldots,\sigma_n\}$, the objective function value of a candidate solution $\pi$ is calculated as follows.
\begin{equation}
    f(\pi)=\frac{1}{n}\sum_{k=1}^{n}d'(\pi,\sigma_k)
\end{equation}
where $d'(\pi,\sigma_k)$ represents the extended Kendall distance between $\pi$ and $\sigma_k$.

To demonstrate the effectiveness of our HER and LADS methods for solving RAP with partial rankings, we experimentally analyze them on benchmark instances and compare it with the extended Borda count method, which operates as follows. Given a set of rankings $\sigma_1,\ldots,\sigma_n$, for each label $\lambda_i$ in a partial ranking of only $m'<m$ labels, if it is a missing label, then it receives $s_{ij}=(m+1)/2$ votes; if it is an existing label with rank $r \in \{1,2,\ldots,m'\}$, then its Borda score is $s_{ij}=(m'+1-r)(m+1)(m'+1)$. The average Borda score $s_i$ is defined as $\frac{1}{n}\sum_{j=1}^{n}s_{ij}$. The labels are then sorted in the decreasing order of their average Borda scores.

To transform a complete ranking into a partial ranking, we resorted to a simple procedure. Given a complete ranking of $n$ items, we execute it from the most to the least preferred item. When item $u$ is visited it can be discarded with a probability $p_d$. If the item is retained, then it stays in the current bucket with probability $p_k$; otherwise, it is randomly assigned to a new bucket. In our experiment, we select $p_d=\frac{2}{3}$ and $p_k=\frac{5}{6}$. There are 20 instances for each combination of $\theta$ and $m$ as well as the complete ranking data. Note that our transformation procedure follows the general practice modeling partial ranking \cite{Aledo2016,Aledo2019}.

\begin{table*}[!hbtp]
\centering
\small
\caption{Comparison of Our Algorithms and Borda on Instances with Partial Rankings}
\label{Tab:Comparison of Our Algorithm and Borda on Instances With Partial Rankings}
\begin{threeparttable}
\setlength{\tabcolsep}{2.7mm}{
\begin{tabular}{lr|r|rrr|rrr}
\toprule[0.75pt]
\multicolumn{2}{c|}{Instance} & \multicolumn{1}{c|}{Borda} & \multicolumn{3}{c}{LADS}& \multicolumn{3}{c}{HER}\\
\midrule[0.5pt]
$\theta$ & $m$ &$f_{best}$ & $f_{best}$ & $f_{avg}$ & $t_{avg}$& $f_{best}$ & $f_{avg}$ & $t_{avg}$\\
\midrule[0.5pt]
0.200&050&110.146 &104.200 &108.439 &201.001 &\textbf{104.010} &\textbf{108.253} &1474.880 \\
0.100&050&161.129 &146.350 &155.746 &259.866 &\textbf{145.710} &\textbf{155.283} &2006.546 \\
0.010&050&260.357 &\textbf{215.040} &227.334 &357.546 &216.440 &\textbf{225.363} &1477.692 \\
0.001&050&271.714 &219.400 &229.024 &380.800 &\textbf{216.920} &\textbf{227.225} &1598.063 \\
0.200&100&322.204 &300.060 &318.294 &2072.679 &\textbf{299.780} &\textbf{318.143} &1920.937 \\
0.100&100&456.761 &\textbf{423.830} &\textbf{444.452} &2818.127 &424.530 &444.579 &2064.247 \\
0.010&100&993.393 &\textbf{865.670} &\textbf{891.501} &3517.686 &875.050 &897.143 &1928.427 \\
0.001&100&1087.184 &901.220 &\textbf{918.750} &3504.064 &\textbf{899.790} &923.105 &1902.893 \\
0.200&150&633.113 &606.310 &629.422 &3562.714 &\textbf{603.630} &\textbf{626.871} &2143.940 \\
0.100&150&847.017 &805.570 &833.089 &3561.592 &\textbf{802.890} &\textbf{829.185} &2367.000 \\
0.010&150&2119.302 &1950.840 &1993.287 &3561.519 &\textbf{1924.550} &\textbf{1964.673} &2533.523 \\
0.001&150&2439.313 &2089.050 &2152.863 &3552.684 &\textbf{2051.680} &\textbf{2099.691} &2911.502 \\
0.200&200&1038.027 &996.220 &1051.096 &3569.241 &\textbf{973.340} &\textbf{1030.689} &2505.553 \\
0.100&200&1322.746 &1273.340 &1323.132 &3572.179 &\textbf{1251.500} &\textbf{1300.955} &2784.889 \\
0.010&200&3573.712 &3381.430 &3463.497 &3568.342 &\textbf{3298.180} &\textbf{3366.975} &3503.892 \\
0.001&200&4330.083 &3877.530 &3940.761 &3559.041 &\textbf{3704.640} &\textbf{3765.796} &3365.421 \\
0.200&250&1512.232 &1524.780 &1574.204 &3576.292 &\textbf{1459.210} &\textbf{1504.668} &3600.000 \\
0.100&250&1887.351 &1855.640 &1922.433 &3575.967 &\textbf{1796.670} &\textbf{1862.538} &3600.000 \\
0.010&250&5281.381 &5074.930 &5209.651 &3534.476 &\textbf{4924.100} &\textbf{5044.982} &3600.000 \\
0.001&250&6735.437 &6157.210 &6261.303 &3519.012 &\textbf{5870.410} &\textbf{5963.002} &3600.000 \\
\midrule[0.5pt]
\#Wins &&20&17&17&$-$&$-$&$-$&$-$\\
\#Ties &&0 &0 &0 &$-$&$-$&$-$&$-$\\
\#Loses&&0 &3 &3 &$-$&$-$&$-$&$-$\\
\midrule[0.5pt]
p-value&& \textbf{8.858e-5} & \textbf{1.300e-3} & \textbf{1.500e-3} &$-$&$-$&$-$&$-$\\
\bottomrule[0.75pt]
\end{tabular}}
\begin{tablenotes}
    \item [$\star$] The result of each combination of $\theta$ and $m$ is averaged over 20 instances.
\end{tablenotes}
\end{threeparttable}
\end{table*}

The comparative results between the proposed algorithms and Borda are summarized in Table \ref{Tab:Comparison of Our Algorithm and Borda on Instances With Partial Rankings}. Note that we run execute each algorithm with a time limit $\hat{t}=1$h. From this table, we observe that our algorithms (i.e., LADS and HER) also show excellent performances on instances with partial rankings. In particular, both the LADS and HER outperform the Borda method for all 20 combinations in terms of both $f_{best}$ and $f_{avg}$. Moreover, the average results of the LADS and HER are better than those achieved by the Borda method. Between HER and LADS, it is not surprising to observe that HER outperforms LADS in terms of $f_{best}$ and $f_{avg}$. This experiment demonstrates the effectiveness of our HER and LADS methods for solving the RAP problem with partial rankings.

It is worth noting that \cite{Aledo2019} proposed and evaluated several basic local search algorithms for RAP with partial rankings on 22 small real-world instances. Unfortunately, their codes are not available to us.

\section{Application to Label Ranking}
\label{Sec:Applied to Label Ranking}

To further demonstrate the practical interest of the proposed ranking aggregation method, we present its application to label ranking (LR), which is an important machine learning task. basically, LR aims to learn a mapping from instances to rankings over a finite number of predefined labels \cite{Hullermeier2008,Cheng2010,Negahban2017,Zhou2018,Alfaro2021}. LR extends the traditional classification and multi-label classification in the view that it must predict the ranking of all class labels rather than only one or several class labels. LR emerges naturally in many areas, such as recommendation systems, image categorization, and meta-learning, \cite{Hullermeier2008,Adomavicius2016,DeSa2017}.

Numerous LR algorithms have been were proposed in the literature owing to their significance \cite{Har2003,Hullermeier2008,Cheng2009,Cheng2010,DeSa2017,Aledo2017,Zhou2018,Alfaro2021}. Decomposition approaches transform the LR problem into several binary classification problems and then combine them into output rankings, such as ranking by pairwise comparison \cite{Hullermeier2008} and constraint classification \cite{Har2003}. Probabilistic approaches represent LR based on statistical models for ranking data, such as instance-based learning algorithms with Mallows \cite{Cheng2009} and Plackett-Luce \cite{Cheng2010} models. In addition to decomposition and probabilistic approaches, ensemble approaches have recently been proposed for solving the LR problem. They usually combine several weak learners to create a more accurate one, such as label ranking forest (LRF) \cite{DeSa2017,Zhou2018} and bagging methods \cite{Aledo2017}. Compared with decomposition and probabilistic approaches, ensemble approaches achieved state-of-the-art performance on LR datasets.

Rank aggregation plays a key role in the LR algorithms. The performance of an LR algorithm depends greatly on the results of the rank aggregation. In an LR algorithm, a set of rankings is usually aggregated by a weak heuristic, that is, the Borda count \cite{Borda1781}. A fast and powerful rank aggregation heurisitc can be used to further improve the existing LR algorithms. To show the interest of our LADS procedure for LR, we integrate LADS into a representative LR algorithm LRF. LRF is an ensemble approach, that obtains state-of-the-art performance on many LR datasets. Moreover, the source code of LRF\footnote{\url{https://github.com/rebelosa/labelrankingforests}} is publicly available \cite{DeSa2017}, which eases our experiments.

\begin{figure}[!htp]
\centering
\includegraphics[width=1.0\columnwidth]{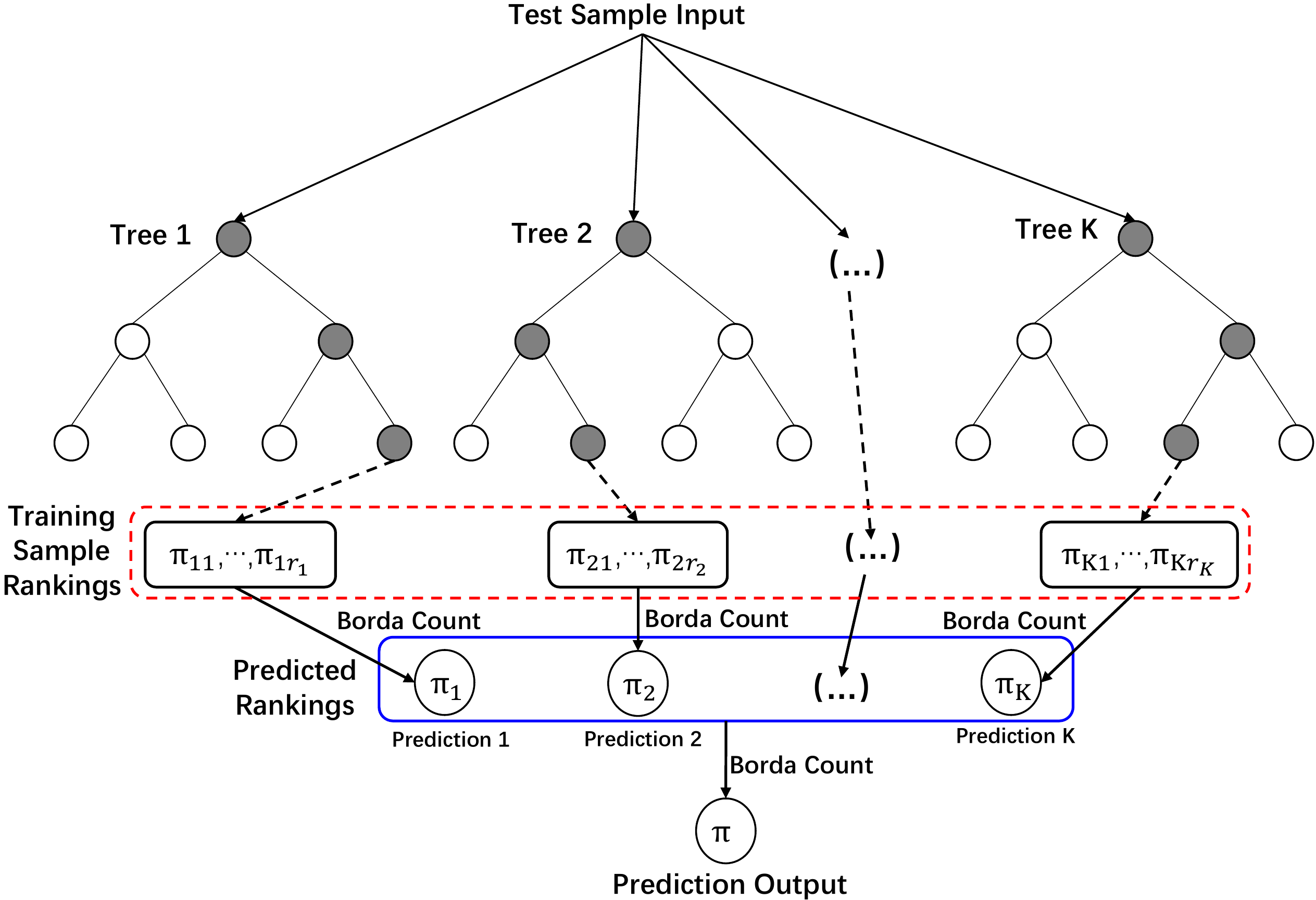}
\caption{Framework of Label Ranking Forest}
\label{Fig:Label Ranking Forest}
\end{figure}

Following the construction of an LRF, it can be used to predict the potential ranking associated with a query sample. Figure~\ref{Fig:Label Ranking Forest} presents the framework of the label ranking forest. During the prediction phase, we pass a test sample through all $K$ trees simultaneously (starting at the root node) until it reaches the leaf nodes. Each decision tree generates a predicted ranking from the target rankings of the training examples in a leaf node. After obtaining $K$ predicted rankings, we aggregate them into a final predicted ranking. Thus, the aggregation of rankings is the rank aggregation problem analyzed in this study. In the prediction phase, LRF requires to perform two types of $K+1$ rank aggregations.
\begin{itemize}
    \item \textbf{Type 1}: Decision tree $i \in \{1,2,\ldots,K\}$ generates a predicted ranking $\pi_i$ based on the rankings of the training samples located in a leaf node, that is, $\{\pi_{i1},\pi_{i2},\ldots,\pi_{ir_i}\}$, where $r_i$ represents the number of training samples in the leaf node.
    \item \textbf{Type 2}: A final predicted ranking $\pi$ is obtained from the $K$ predicted rankings generated by $K$ decision trees, that is, $\{\pi_1,\pi_2,\ldots,\pi_K\}$.
\end{itemize}

To demonstrate the usefulness of our rank aggregation method to enhance the standard LRF approach, we use the LADS algorithm to perform the rank aggregation task of LRF and compare the standard LRF approach with three LRF variants. Specifically, LRF$_{10}$ is obtained from the LRF by only performing rank aggregation of type 1 with LADS; LRF$_{01}$ presents a variant of LRF by only performing rank aggregation of type 2 with LADS; LRF$_{11}$ is a variant of LRF by performing rank aggregation of both type 1 and type 2 with LADS.

Our experiments are conducted on six semi-synthetic and three real-world datasets randomly selected from widely used LR datasets\footnote{\url{https://en.cs.uni-paderborn.de/de/is/research/research-projects/software/label-ranking-datasets}}. Following general practice \cite{Hullermeier2008,Cheng2010,Zhou2018}, we use Kendall's tau coefficient \cite{Kendall1938} to evaluate the performance of LR algorithms. We construct a label ranking forest of $K=100$ decision trees and use the default parameters in our experiments. Table \ref{Tab:Comparisons on Semi-synthetic Datasets} summarizes the comparative results of the LRF and the three variants enhanced by our LADS algorithm on semi-synthetic datasets. At its bottom, we also provide the average rank of each algorithm for all tested instances. We first order the algorithms according to their performances, and average ranks are assigned in the case of ties. For the indicator of average rank, the smaller the value, the better the algorithm.

\begin{table*}[!hbtp]
\small
\caption{Comparison of LRF and its Variants on Semi-synthetic Datasets for Label Ranking}
\label{Tab:Comparisons on Semi-synthetic Datasets}
\begin{center}
\begin{threeparttable}
\setlength{\tabcolsep}{3.5mm}{
\begin{tabular}{l|crr|rrrr}
\toprule[0.75pt]
\multicolumn{1}{c|}{Data sets} & \multicolumn{1}{c}{\#Samples} & \multicolumn{1}{c}{\#Features} & \multicolumn{1}{c|}{\#Labels} & \multicolumn{1}{c}{LRF} & \multicolumn{1}{c}{LRF$_{10}$}& \multicolumn{1}{c}{LRF$_{01}$} & \multicolumn{1}{c}{LRF$_{11}$}\\
\midrule[0.5pt]
authorship  &841&70&4&0.892 &\textbf{0.893} &\textbf{0.892} &\textbf{0.892}\\
bodyfat&252&7 &7 &0.203 &0.200 &\textbf{0.206} &\textbf{0.207}\\
glass  &214&9 &6 &0.885 &\textbf{0.893} &\textbf{0.887} &\textbf{0.894}\\
housing&506&6 &6 &0.804 &\textbf{0.809} &\textbf{0.807} &\textbf{0.811}\\
iris   &150&4 &3 &0.956 &\textbf{0.956} &\textbf{0.959} &\textbf{0.960}\\
vehicle & 846 & 18 & 4 & 0.860 & \textbf{0.861} & \textbf{0.860} & \textbf{0.862} \\
\midrule[0.5pt]
avg. rank & $-$ & $-$ & $-$ & 3.583 & 2.333 & 2.750 & \textbf{1.333}\\
\bottomrule[0.75pt]
\end{tabular}}
\begin{tablenotes}
    \item [$\star$] The results are obtained using a four-fold cross validation.
\end{tablenotes}
\end{threeparttable}
\end{center}
\end{table*}

From Table \ref{Tab:Comparisons on Semi-synthetic Datasets}, we observe that the three variants of LRF obtain smaller average ranks than LRF, indicating that LADS can significantly improve LRF. Specifically, LRF$_{10}$ achieves better or equal results on five out of six tested datasets, and worse performance on one dateset. LRF$_{01}$ obtains better performance on four out of six tested datasets, and the same performance on two datasets. LRF$_{11}$ achieves better results on five out of six tested datasets, and the same result on one dataset. This experiment demonstrates the interest of the LADS procedure for the important label ranking problem.

Table \ref{Tab:Comparisons on Real-world Datasets} lists the comparative results of LRF and its three variants on real-world datasets. From this table, we also observe that LADS can significantly enhance LRF. The variants of LRF outperforms the original LRF in terms of the average rank. In particular, LRF$_{11}$ obtains the smallest average rank 1.0, LRF$_{01}$ obtains the second-best average rank 2.166. These observations confirm the benefit of our LADS algorithm to improve existing label ranking algorithms.

\begin{table*}[!hbtp]
\small
\caption{Comparison of LRF with its Variants on Real-world Datasets for Label Ranking}
\label{Tab:Comparisons on Real-world Datasets}
\begin{center}
\begin{threeparttable}
\setlength{\tabcolsep}{3.5mm}{
\begin{tabular}{l|crr|rrrr}
\toprule[0.75pt]
\multicolumn{1}{c|}{Data sets} & \multicolumn{1}{c}{\#Samples} & \multicolumn{1}{c}{\#Features} & \multicolumn{1}{c|}{\#Labels} & \multicolumn{1}{c}{LRF} & \multicolumn{1}{c}{LRF$_{10}$}& \multicolumn{1}{c}{LRF$_{01}$} & \multicolumn{1}{c}{LRF$_{11}$}\\
\midrule[0.5pt]
cold&2465&24&4 &0.076 &\textbf{0.078} &\textbf{0.081} &\textbf{0.086}\\
dtt &2465&24&4 &0.114 &0.110 &\textbf{0.118} &\textbf{0.121}\\
heat&2465&24&6 &0.028 &\textbf{0.029} &\textbf{0.029} &\textbf{0.030}\\
\midrule[0.5pt]
avg. rank & $-$ & $-$ & $-$ & 3.667 & 3.167 & 2.166 & \textbf{1.000}\\
\bottomrule[0.75pt]
\end{tabular}}
\begin{tablenotes}
    \item [$\star$] The results are obtained using a four-fold cross validation.
\end{tablenotes}
\end{threeparttable}
\end{center}
\end{table*}

\section{Analysis and Discussion}
\label{Sec:Discussion and Analysis}

In this section, we present additional experiments to gain a deeper understanding of HER. We perform three groups of experiments: 1) to study the effect of $L_h$, 2) to investigate the benefit of the incremental evaluation technique, and 3) to evaluate the effectiveness of CPSC. The following experiments were conducted on 10 representative instances, where each instance is selected based on its $\theta$ and $m$ values.

\subsection{Effect of $L_h$}
\label{SubSec:Effect of the History Length on LADS}

\begin{figure}[!htp]
\centering
\includegraphics[width=1.0\columnwidth]{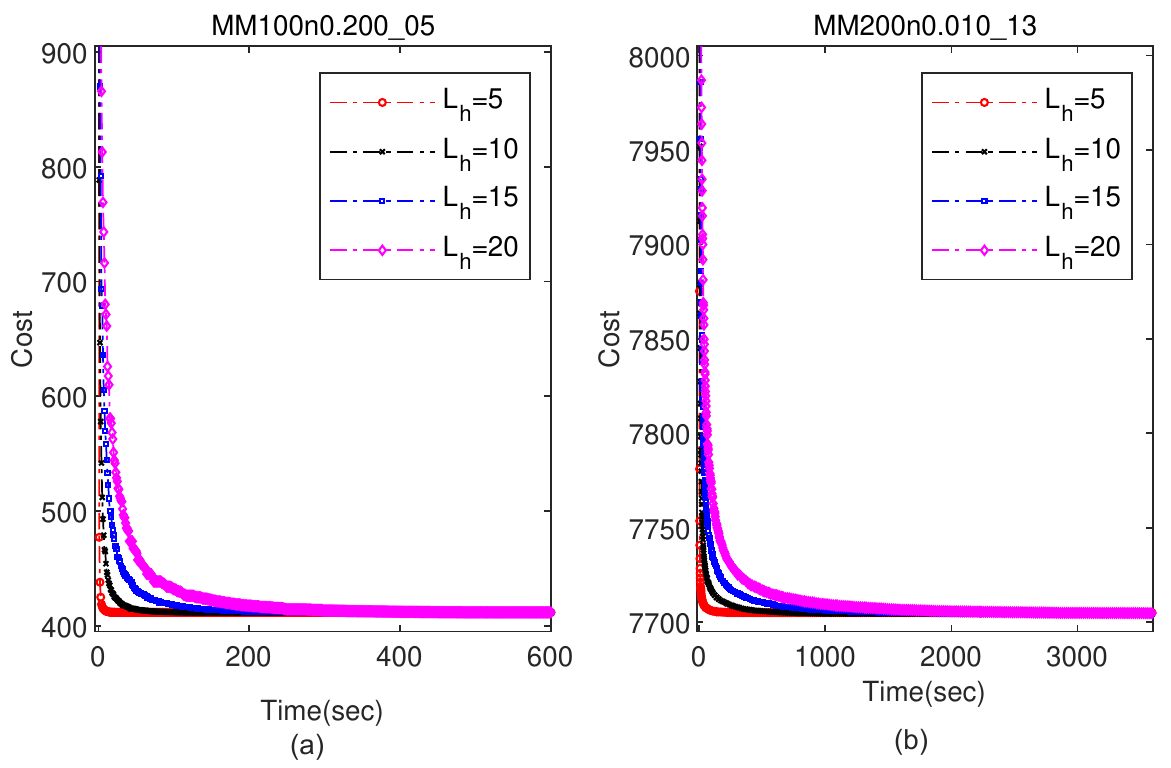}
\caption{The Cost Drop Diagrams of LADS with Different $L_h$ Values}
\label{Fig:Running Profiles}
\end{figure}

The history length $L_h$ is an important parameter of LADS, which determines the convergence speed and solution quality. To study the effect of $L_h$, we test different $L_h$ values from 5 to 20 with a step size of 5. Figure~\ref{Fig:Running Profiles} depicts the comparative performance of LADS with these $L_h$ values on two instances MM100n0.200\_05 and MM200n0.010\_13.

From Figure~\ref{Fig:Running Profiles}(a), we can observe that starting from the same initial solution, LADS with $L_h=5$ rapidly improves the cost. LADS with $L_h=10$ improves the cost more slowly, with $L_h=15$ even more slowly, and $L_h=20$ yields the slowest improvement in the cost. In other words, the larger the $L_h$, the slower the cost decreases. From Figure~\ref{Fig:Running Profiles}(b), we can also observe the same. It should be noted that the cost drop diagrams plotted for the other eight selected instances are similar to these two instances. To balance the solution quality and convergence speed, we determine $L_h=5$ in our LADS algorithm, which achieves the fastest (over the four curves) improvement of the cost function.

\subsection{Benefit of Incremental Evaluation Technique}
\label{SubSec:Benefit of Incremental Evaluation Technique}

\begin{figure}[!htp]
\centering
\includegraphics[width=1.0\columnwidth]{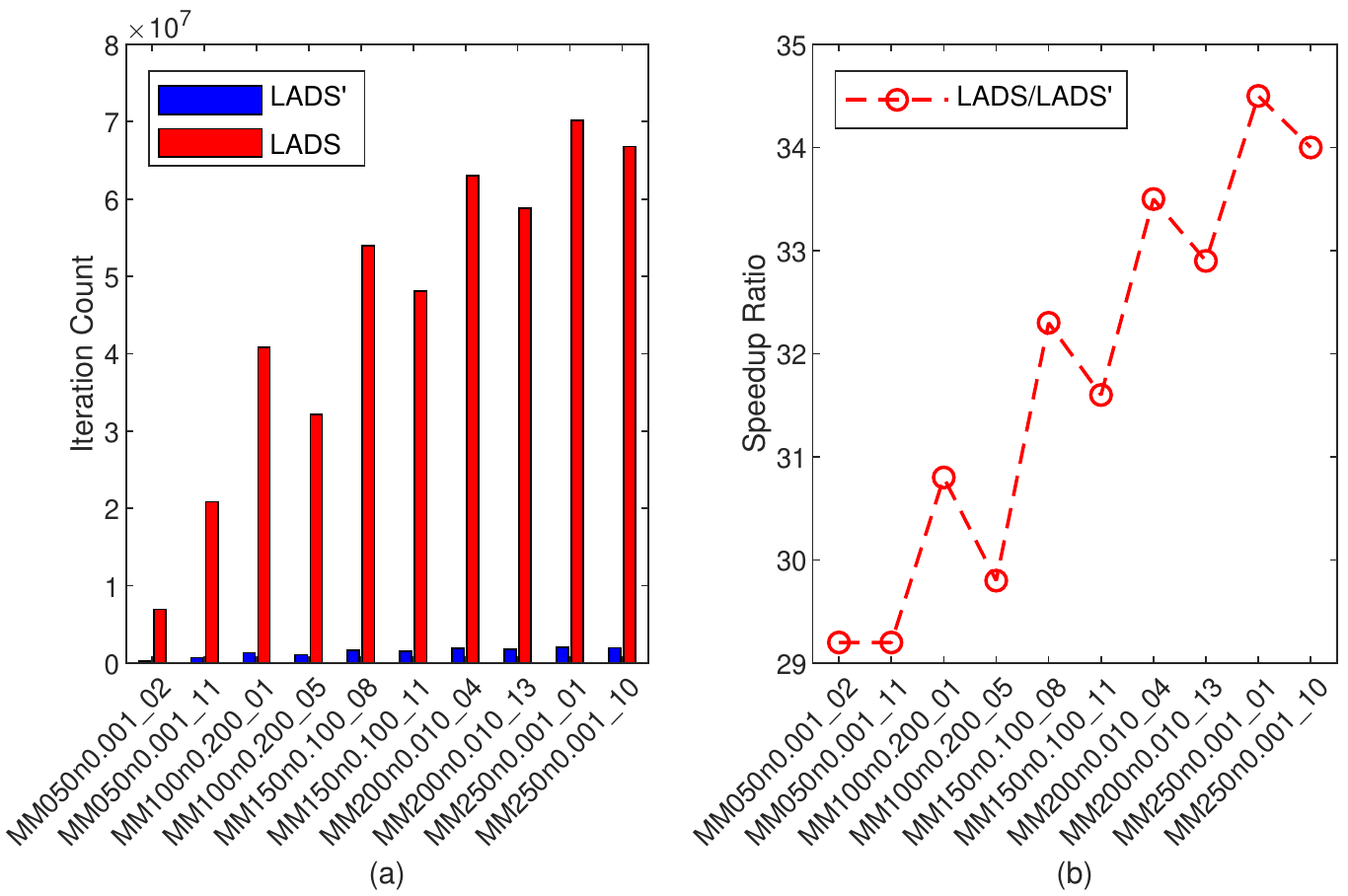}
\caption{Comparison between LADS with and without the Incremental Evaluation Technique.}
\label{Fig:Speedup Ratio}
\end{figure}

To demonstrate the benefit of the incremental evaluation technique, we experimentally compare the LADS with a variant LADS$'$ where the incremental evaluation technique is disabled. For each algorithm, we execute it on the 10 selected instances with the limit time $\hat{t}=100$ s and record the total number of iterations. For both LADS and LADS$'$, we set their history lengths $L_h=5$ according to Section \ref{SubSec:Effect of the History Length on LADS}. The comparative results of these two algorithms are summarized in Figure~\ref{Fig:Speedup Ratio}.

Figure~\ref{Fig:Speedup Ratio}(a) shows the comparative performance between LADS and LADS$'$ in terms of the total number of iterations (i.e., iteration count). From this figure, we observe that LADS can perform more iterations than LADS$'$ for a given time. This allows LADS to sample more candidate solutions and thus increases its chance of finding solutions of better quality. Figure~\ref{Fig:Speedup Ratio}(b) shows the speedup ratio of the LADS over LADS$'$, which is approximately 29 to 34 for each instance. These observations demonstrate the benefits of the incremental evaluation technique for the search algorithm.

\subsection{Effectiveness of Concordant Pairs-based Semantic Crossover}
\label{SubSec:Effectiveness of Concordant Pairs-based Semantic Crossover}

To demonstrate the effectiveness of CPSC, we experimentally compare HER with three variants, namely HER$'$, HER$''$, and HER$'''$ where CPSC is replaced by three other popular permutation crossover operators, i.e., order crossover, order-based crossover, and position-based crossover \cite{Pavai2016}, respectively.

\begin{table*}[!hbtp]
\small
\caption{Comparison of HER with the CPSC Operator and its Three Variants with Other Crossovers}
\label{Tab:Comparison Between HER and Different Variants}
\begin{center}
\begin{threeparttable}
\resizebox{\textwidth}{!}{
\begin{tabular}{c|rr|rr|rr|rr}
\toprule[0.75pt]
\multicolumn{1}{c|}{}  & \multicolumn{2}{c|}{HER$'$} & \multicolumn{2}{c|}{HER$''$}& \multicolumn{2}{c|}{HER$'''$} & \multicolumn{2}{c}{HER}\\
\midrule[0.5pt]
Instance &$f_{best}$ & $f_{avg}$ & $f_{best}$ & $f_{avg}$& $f_{best}$ & $f_{avg}$& $f_{best}$ & $f_{avg}$\\
\midrule[0.5pt]
MM050n0.001$\_$02&\textbf{575.480} &575.516 &\textbf{575.480} &575.534 &\textbf{575.480} &575.506 &\textbf{575.480} &\textbf{575.502} \\
MM050n0.001$\_$11&\textbf{561.480} &\textbf{561.480} &\textbf{561.480} &561.486 &\textbf{561.480} &\textbf{561.480} &\textbf{561.480} &\textbf{561.480} \\
MM100n0.200$\_$05&\textbf{411.720} &\textbf{411.720} &\textbf{411.720} &\textbf{411.720} &\textbf{411.720} &\textbf{411.720} &\textbf{411.720} &\textbf{411.720} \\
MM100n0.200$\_$01&\textbf{416.000} &\textbf{416.000} &\textbf{416.000} &\textbf{416.000} &\textbf{416.000} &\textbf{416.000} &\textbf{416.000} &\textbf{416.000} \\
MM150n0.100$\_$17&\textbf{1266.090} &\textbf{1266.100} &\textbf{1266.090} &1266.114 &\textbf{1266.090} &\textbf{1266.100} &\textbf{1266.090} &\textbf{1266.100} \\
MM150n0.100$\_$08&\textbf{1249.580} &\textbf{1249.598} &\textbf{1249.580} &1249.622 &\textbf{1249.580} &\textbf{1249.598} &\textbf{1249.580} &\textbf{1249.598} \\
MM200n0.010$\_$13&7704.840 &7704.978 &7704.920 &7705.284 &\textbf{7704.800} &\textbf{7704.938} &\textbf{7704.800} &7705.060 \\
MM200n0.010$\_$04&\textbf{7667.850} &\textbf{7668.052} &7668.030 &7668.366 &\textbf{7667.850} &7668.056 &7667.910 &7668.054 \\
MM250n0.001$\_$10&14442.240 &14442.674 &14442.340 &14443.086 &14442.540 &14442.964 &\textbf{14442.140} &\textbf{14442.604} \\
MM250n0.001$\_$01&\textbf{14353.330} &14353.888 &14353.870 &14354.682 &14353.490 &14353.912 &\textbf{14353.330} &\textbf{14353.650} \\
\midrule[0.5pt]
avg.value&4864.861 &4865.001 &4864.951 &4865.189 &4864.903 &4865.027 &\textbf{4864.853} &\textbf{4864.977} \\
\midrule[0.5pt]
avg.rank&2.300 &2.100 &3.000 &3.700 &2.500 &2.300 &\textbf{2.200} &\textbf{1.900} \\
\bottomrule[0.75pt]
\end{tabular}}
\end{threeparttable}
\end{center}
\end{table*}

Table \ref{Tab:Comparison Between HER and Different Variants} summarizes the comparative results of the HER and its three variants on the 10 selected instances. In this table, we report the best result $f_{best}$ and the average result $f_{avg}$ of each algorithm over ten runs. We also list the average value at the end of each column, and the average rank of each performance indicator. We observe that HER outperforms all the variants, achieving a better average value and average rank in terms of both $f_{best}$ and $f_{avg}$. These observations confirm the effectiveness of the CPSC used in HER.

\section{Concluding Remarks}
\label{Sec:Concluding Remarks}

In this paper, we proposed an effective hybrid evolutionary ranking algorithm for solving the challenging rank aggregation problem with both complete and partial rankings. To generate promising offspring solutions, the algorithm uses a problem-specific crossover based on concordant pairs between two parent solutions. Moreover, the algorithm integrates a powerful local optimization procedure combining the late acceptance strategy and a fast incremental evaluation technique introduced for the first time in this study. Empirical results on various benchmark instances of both complete and partial rankings showed excellent performance of the proposed method compared to the existing methods.

To further demonstrate the usefulness of the proposed method for practical problems, we applied our method to label ranking, which is a relevant task in machine learning. This study showed that our ranking method can benefit label ranking algorithms by generating better rank aggregations.

There are several perspectives for future research. First, it would be interesting to test the proposed method for other applications. The codes of the proposed algorithms that are publicly available facilitate such applications. Second, the incremental evaluation technique introduced in this study is general and can benefit other search-based algorithms for the rank aggregation problem and boost their computational efficiency. Third, a concordant pairs-based semantic crossover is designed for permutation encoding and enriches the pool of existing permutation crossovers. Thus, this crossover may find interesting applications in which an order relation among the permutation elements is relevant. Finally, in recent years, many efforts have been made to use machine learning techniques to improve optimization methods. This work contributes to the research on the use of optimization methods to solve machine learning problems more efficiently.


\section*{Acknowledgment}
We would like to thank Zhihao Wu and Jiaoyan Guan for helping to perform some experiments.

\ifCLASSOPTIONcaptionsoff
  \newpage
\fi

\bibliographystyle{IEEEtran}
\bibliography{mybibfiles}

\end{document}